\pdfoutput=1

\documentclass[11pt]{article}

\usepackage{ACL2023}

\usepackage{times}
\usepackage{latexsym}

\usepackage[T1]{fontenc}

\usepackage[utf8]{inputenc}

\usepackage{microtype}

\usepackage{graphicx}
\usepackage{acronym}
\usepackage[inline]{enumitem}
\usepackage[normalem]{ulem}
\usepackage{caption}
\usepackage{multirow}
\usepackage{cite}
\usepackage{natbib}
\usepackage{multirow}
\usepackage{booktabs}
\usepackage{amsmath}
\usepackage{xcolor}
\usepackage{amsfonts}
\usepackage{amssymb}
\usepackage{pifont}
\usepackage[toc,page]{appendix}

\newcommand{\changed}[1]{#1}

\newcommand{\code}[1]{{\ttfamily#1}}

\newcommand{\header}[1]{\vspace*{1mm}\noindent\textbf{#1.}}

\title{Answering Ambiguous Questions via Iterative Prompting}

\author{
Weiwei Sun\textsuperscript{\rm 1} \quad Hengyi Cai\textsuperscript{\rm 2} \quad Hongshen Chen\textsuperscript{\rm 2} \quad Pengjie Ren\textsuperscript{\rm 1}
\\ 
{\bf Zhumin Chen\textsuperscript{\rm 1} \quad  Maarten de Rijke\textsuperscript{\rm 3} \quad Zhaochun Ren\textsuperscript{\rm 1}\thanks{~~Corresponding author.}}
\\
\textsuperscript{\rm 1}Shandong University, Qingdao, China\\
\textsuperscript{\rm 2}JD.com, Beijing, China\\
\textsuperscript{\rm 3}University of Amsterdam, Amsterdam, The Netherlands\\
\texttt{\{sunnweiwei,hengyi1995\}@gmail.com}\\
\texttt{ac@chenhongshen.com,m.derijke@uva.nl}\\
\texttt{\{renpengjie,chenzhumin,zhaochun.ren\}@sdu.edu.cn}
}

\begin{document}
\maketitle
\begin{abstract}



In open-domain question answering, due to the ambiguity of questions, multiple plausible answers may exist.
To provide feasible answers to an ambiguous question,
one approach is to directly predict all valid answers, but this can struggle with balancing relevance and diversity.
An alternative is to gather candidate answers and aggregate them, but this method can be computationally costly and may neglect dependencies among answers.
%
In this paper, we present \emph{AmbigPrompt} to address the imperfections of existing approaches to answering ambiguous questions.
Specifically, we integrate an answering model with a prompting model in an iterative manner.
The prompting model adaptively tracks the reading process and progressively triggers the answering model to compose distinct and relevant answers. 
Additionally, we develop a task-specific post-pretraining approach for both the answering model and the prompting model, which greatly improves the performance of our framework. 
Empirical studies on two commonly-used open benchmarks show that AmbigPrompt achieves state-of-the-art or competitive results while using less memory and having a lower inference latency than competing approaches. 
Additionally, AmbigPrompt also performs well in low-resource settings. 
The code are available at: \url{https://github.com/sunnweiwei/AmbigPrompt}.
\end{abstract}

\section{Introduction}


Recent years have witnessed substantial advances in open-domain question answering (QA) systems~\citep{Karpukhin2020DensePR,Lewis2020RetrievalAugmentedGF,Izacard2021LeveragingPR}, which aim to find the answer for the given question from a large knowledge corpus~\citep{Chen2017ReadingWT}.
While a dominating scenario is the single-answer QA setting, i.e., only one exact answer is required for a given question~\citep{Karpukhin2020DensePR}, 
this work focuses on the more realistic scenario of \emph{Multi-answer QA}, where multiple plausible answers are associated with a user-issued question~\citep{Min2020AmbigQAAA}, given that questions posed by humans are often open-ended and ambiguous.\footnote{The task of this paper primarily focuses on the occurrence of multiple answers resulting from different interpretations caused by question ambiguity.
However, it's worth to note that question ambiguity is just one factor contributing to the presence of multiple answers. In this study, we adhere to the conceptual definition of \citet{Min2020AmbigQAAA}.}

\begin{figure}[!t]
 \centering
 \includegraphics[width=1\columnwidth]{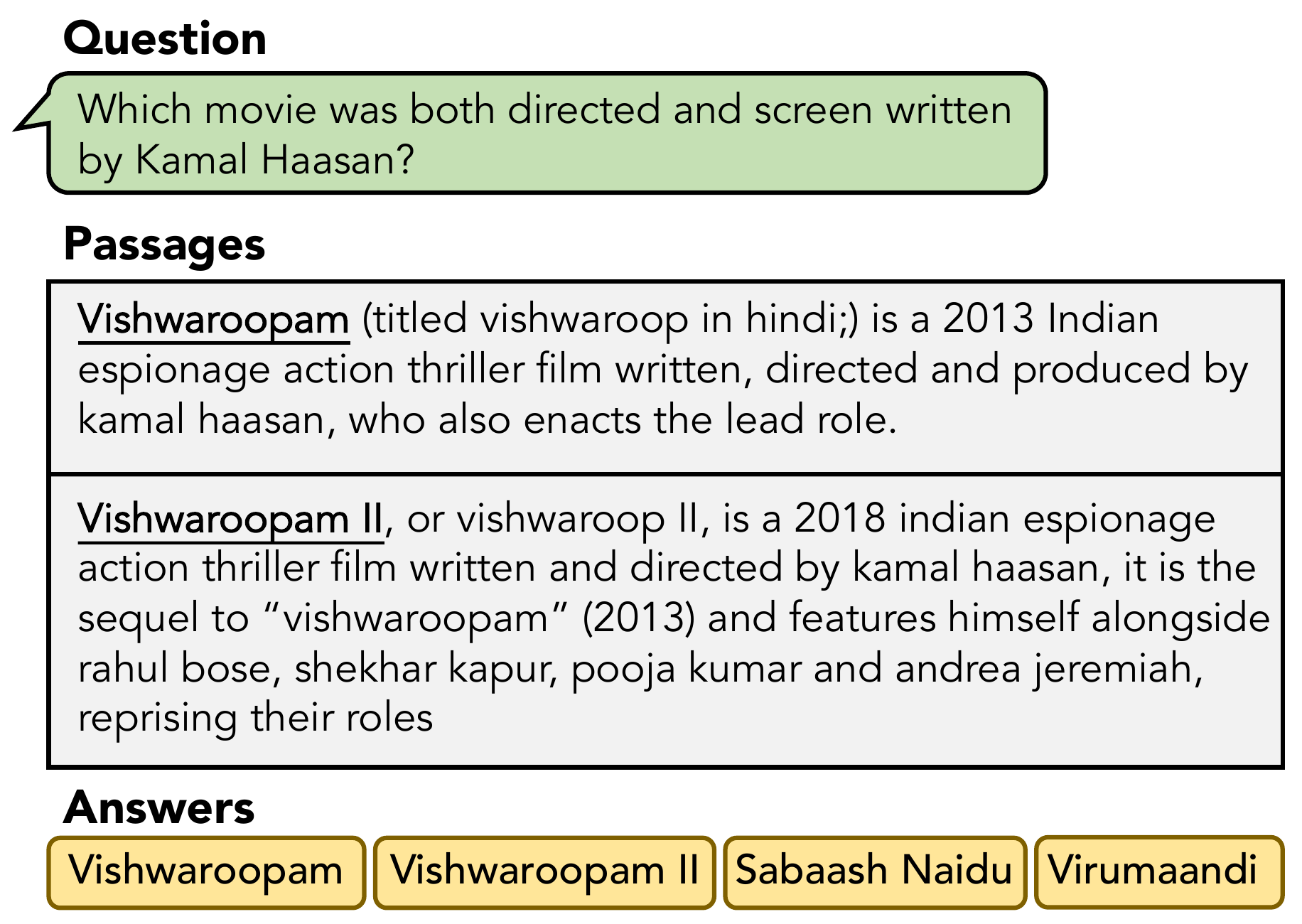}
 \caption{An example of an open-domain question, a subset of its evidential Wikipedia passages and multiple answers they lead to.}
 \label{fig:example}
\end{figure}

A natural approach for answering ambiguous open-domain questions would be to fine-tune a pre-trained answer generation model, e.g., T5~\citep{Raffel2020ExploringTL}, using supervised data of the form (evidential passages, question, all plausible answers)~\citep{Min2020AmbigQAAA,Min2021JointPR}.
However, this approach often leads to sub-optimal solutions 
since it requires the model to balance the relevance and diversity of the generated multiple answers within a single-round decoding procedure, which is non-trivial.
To manage the relevance-diversity trade-off, another approach is to decompose multi-answer QA into candidate answer prediction and answer post-processing. 
This typically requires a high-capacity model with billions of parameters to construct candidate answers and sophisticated answer aggregation pipelines to obtain the final results~\citep{Shao2021AnsweringOM,Gao2021AnsweringAQ}, incurring high computational costs.
In addition, this approach suffers from the dilemma of having to predict diverse candidate answers before knowing which answer has been predicted, which is unnatural and intricate.
For example, in Figure~\ref{fig:example}, given the question ``\emph{Which movie was both directed and screenwritten by Kamal Haasan?},'' with the existence of the answer \emph{Vishwaroopam}, 
the model excludes its eponymous translation version \emph{Vishwaroop} and deduces that \emph{Vishwaroopam II} is another potential answer.

\begin{figure}[!t]
 \centering
 \includegraphics[width=1\columnwidth]{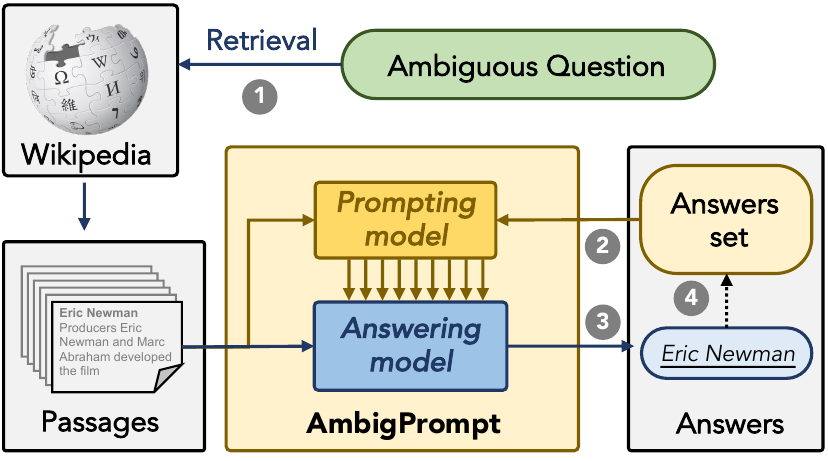}
 \caption{Given the retrieved passages, AmbigPrompt alternates between (2) generating prompts based on previous answers, (3) generating a new answer using a question-answering model, and (4) appending the new answer to the answers set. Note that steps (2) and (3) operate in an interleaving way.}
 \label{fig:model}
\end{figure}

When facing an ambiguous question, people are capable of providing multiple valid answers by introspectively composing new content on the basis of what has already been devised, usually in an iterative manner.
Inspired by this observation, in this paper, we conceptualize \textbf{AmbigPrompt} as an approach to mimic this mechanism by iteratively guiding the answering model with a lightweight prompting model.
As shown in Figure~\ref{fig:model}, this prompting model steers the answering model to progressively generate valid answers whose content the prompting model will then condition on for the next-round prompt construction.
Essentially, our proposed framework comprises two key components: (i) an encoder-decoder \emph{answering model} and (ii) an interleaving answer-conditional \emph{prompting model}.
By conditioning on preceding generated contents,
the proposed framework introspectively perceives which answer has been predicted before updating the hidden activation for the generation of subsequent answers.
Furthermore, we devise a task-adaptive post-pretraining strategy,
in which pseudo multi-QA training instances are constructed to facilitate the training of the proposed framework.


We carry out extensive experiments on the AmbigQA~\citep{Min2020AmbigQAAA} and WebQSP~\citep{Yih2016TheVO} datasets. 
The results demonstrate that AmbigPrompt attains superior performance despite having a significantly smaller parameter scale, 14 times less than state-of-the-art models.
Furthermore, as a lightweight approach, AmbigPrompt improves the answer relevance and diversity with a tiny fraction of the memory footprint and inference latency of competing approaches.
Notably, AmbigPrompt achieves the best performance in the low-resource setting.
The effectiveness of the proposed method is also verified by ablation experiments and analytical experiments.

In summary, this paper makes the following contributions:
\begin{enumerate*}[label=(\roman*)]
    \item We propose AmbigPrompt, which tackles ambiguous question answering by iterative prompting.
    \item We propose an interleaving answer-conditional prompting model to generate meaningful continuous prompts.
    \item Experiments on multi-QA datasets verify the effectiveness of the proposed approach.
\end{enumerate*}


\section{Preliminaries} \label{preliminary}

\subsection{Problem formalization}

Formally, given an open-domain question $q$, a multi-answer question answering (QA) model is required to make use of (multiple pieces of) evidence from a large-scale text corpus $\Omega$ (e.g., Wikipedia)
to find multiple plausible answers $\mathcal{A} = \{a_1,a_2,\ldots,a_n\}$, where $a_i$ denotes one answer and we suppose there are $n$ answers.
The QA model aims to infer $p(\mathcal{A}|q,\Omega)$.
In open-domain QA, the QA model typically follows a two-step pipeline, comprising \emph{passage retrieval} and \emph{answer generation}.
In the passage retrieval step, a retrieval model $p(\mathcal{C}|q,\Omega)$ retrieves $m$ evidence passages $\mathcal{C} = \{c_1,c_2,\ldots,c_m\}$ according to the question $q$ from $\Omega$.
In the answer generation step, an answering model $p(\mathcal{A}|q,\mathcal{C})$ reads the evidential passages and finds the answers to the question.

\subsection{Answering model} 
We use Fusion-in-Decoder (FiD) as a basic single-answer answering model~\citep{Izacard2021LeveragingPR}.
In particular, FiD has an encoder-decoder architecture.
FiD first concatenates each retrieved passage with the question with a \code{[SEP]} token:
\begin{equation}\label{eq:context}
    X=\{x_1,x_2,\ldots,x_m\}, 
    \
    x_i = q\text{\code{[SEP]}}c_i
\end{equation}
where we use $X$ to denote the concatenated sequence.
Then, for each $x_i$, the FiD encoder $\operatorname{Enc}$ encodes it to $\mathbf{x}_i$:
\begin{equation}
    \mathbf{X}=\operatorname{Cat}(\{\mathbf{x}_1, \mathbf{x}_2,\ldots,\mathbf{x}_m\}), 
    \ 
    \mathbf{x}_i = \operatorname{Enc}(x_i)
\end{equation}
where $\operatorname{Cat}$ denotes a concatenation function.
Finally, the decoder $\operatorname{Dec}$ attends to the representations of all passages and generates an answer $a$:
\begin{equation}\label{eq:fid-dec}
    p(a|q, \mathcal{C})=\operatorname{Dec}(\mathbf{X})
\end{equation}

\subsection{Prompt-tuning}
Prompt-tuning adapts pre-trained transformer models to downstream tasks by optimizing continuous prompting vectors~\citep{Li2021PrefixTuningOC,Liu2021PTuningVP}.
Suppose $x$ is the input sequence of the model,
we denote $Q(x)^{j}$, $K(x)^{j}$, $V(x)^{j}$ as the query, key, and value representations of $x$ in the $j$-th attention layer in the transformer encoder.
Prompt-tuning prepends learnable prompting vectors $\mathbf{E}^j$ to $K(x)^j$ and $V(x)^j$ to modify the attention distribution as well as the output $\mathbf{x}^j$ of the $j$-th layer as follows:
\begin{equation}
\begin{split}
\mathbf{x}^j = \operatorname{Attn}(Q(x)^j, &\operatorname{Cat}(\mathbf{E}^j,K(x)^j),\\
&\operatorname{Cat}(\mathbf{E}^j,V(x)^j)),
\end{split}
\label{eq:prompt-tuning}
\end{equation}
where $\mathbf{x}^j$ denotes the output of layer $j$, $\operatorname{Attn}(\cdot)$ represents the attention operation in the transformer, and $\operatorname{Cat}(\cdot)$ is the concatenation function.

\section{AmbigPrompt}

Conventionally, the question answering model generates the desired answer given the input context in a single pass~\citep{Izacard2021LeveragingPR}.
While it suffices to tackle the single-answer QA scenario, managing ambiguous questions with multiple answers can be more nuanced -- 
the answering model is required to balance the relevance and diversity of the generated answers in a single pass, and precisely modeling dependencies among the answers can be non-trivial.
In this paper, we propose AmbigPrompt, a question-answering model that answers ambiguous questions via iterative prompting, inferring more accurate answers progressively. 
Figure \ref{fig:model} gives an overview of the proposed method.

Overall, AmbigPrompt decomposes the generation of answers $\mathcal{A}$ into multiple steps instead of one single pass, i.e.,
\begin{equation} \label{eq:iter}
    p(\mathcal{A}|q,\mathcal{C})= \prod_{t=1}^n p(a_t|\phi(a_{<t}),q,\mathcal{C}),
\end{equation}
where $a_{<t}$ denotes the set of answers that have been generated at time $t$, and $\phi(\cdot)$ denotes a prompting model that generates prompt vectors for answer generation at the $t$-th step.
\changed{The prompting model shares parameters with the answering model, allowing for seamless integration.}
AmbigPrompt iteratively composes a new answer $a_t$, conceiving the prompt of previous answers, i.e., $\phi(a_{<t})$, and appends $a_t$ to the answers set, till all feasible answers are found.

The proposed framework is optimized in a two-stage manner: \emph{task-adaptive post-pretraining} and \emph{prompt-based tuning}.
In the former stage, the model is trained on a large synthesized multi-answer QA dataset, 
while in the latter stage, the model is tuned on the annotated multi-answer QA dataset.
We first detail the prompting model (\S\ref{sec:prompt}) and the iterative question answering procedure (\S\ref{sec:iterative-prompt}), and then introduce the optimization scheme (\S\ref{sec:optimization}).


\begin{figure*}[!t]
 \centering
 \includegraphics[width=2\columnwidth]{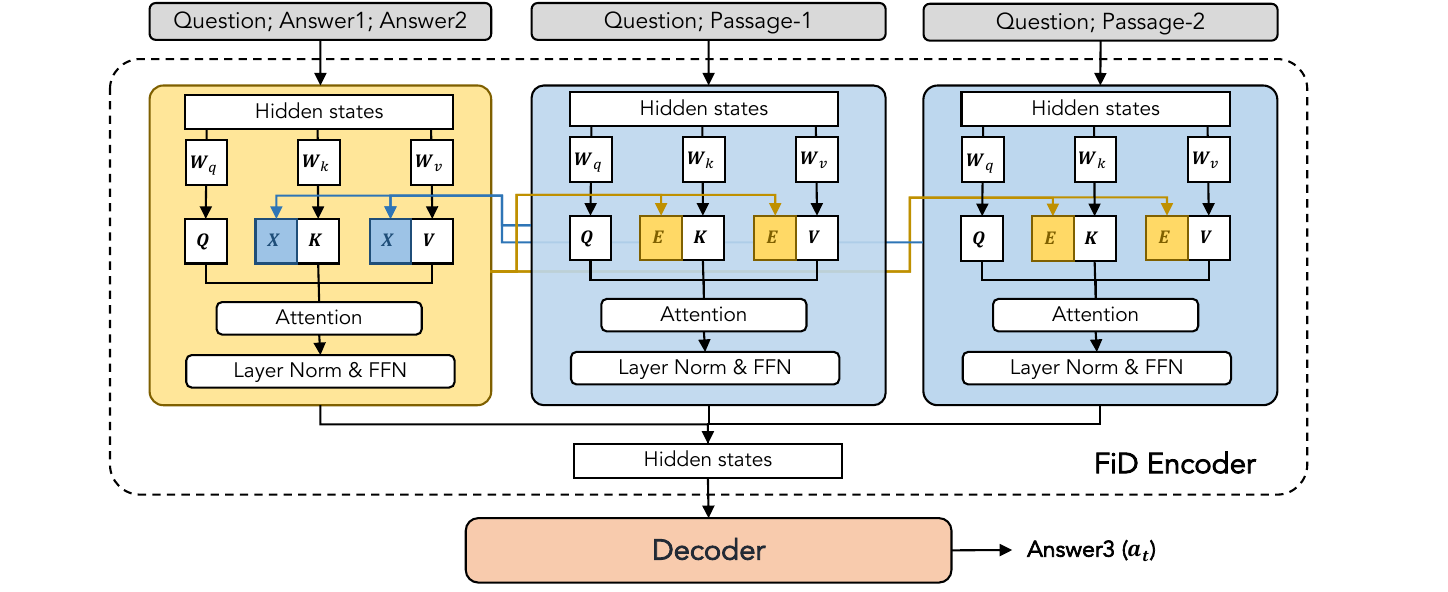}
 \caption{Details of the retrospective prompting mechanism. The prompting model produces the prompt vectors $\mathbf{E}$ by cross-attending the contextual representation $\mathbf{X}$. And the answering model predicts a new answer $a_t$ using the prompt $\mathbf{E}$. The prompting and answering models operate in an interleaving manner.}
 \label{fig:opt}
\end{figure*}

\subsection{Retrospective prompting mechanism for answer generation} 
\label{sec:prompt}
To capture intricate dependencies among answers, we devise an interleaving answer-conditional prompting model $\phi(a_{<t})$, which generates the prompt vector $\mathbf{E}=\phi(a_{<t})$ conditioned on antecedent generated answers $a_{<t}$, as depicted in Figure~\ref{fig:opt}.
Specifically, the prompting model $\phi$ is a transformer encoder that shares the same parameters with the encoder of the answering model. $\phi$ processes the $a_{<t}$ in three steps:
\begin{enumerate}[label=(\arabic*),leftmargin=*]
    \item \textbf{Templating answers.} First, $a_{<t}$ is transformed into a text sequence $e=\mathcal{T}(a_{<t})$ using a template $\mathcal{T}$. Here we use semicolons to splice answers.
    
    \item  \textbf{Generating prompts.} 
    Then, given the answer sequence $e$ and context $X$ \changed{(i.e., the concatenated question and passages in Eq.~\ref{eq:context})}, the prompting model $\phi$ computes the hidden activations $\mathbf{E}^j$ of each layer $j$ via cross-attending the contextual representation $\mathbf{X}^{j-1}$:
    \begin{equation}
    \begin{split}
    \mathbf{E}^j=\operatorname{Attn}(Q(e)^j,&\operatorname{Cat}(K(e)^j, \mathbf{X}^{j-1}),\\
    &\operatorname{Cat}(V(e)^j, \mathbf{X}^{j-1})),
    \end{split}
    \end{equation}
    where $Q(e)^j$, $K(e)^j$, and $V(e)^j$ denote the query, key, and value representations of $e$ in the $j$-th attention layer in the prompting model; $\mathbf{X}^{j-1}{=}\operatorname{Cat}(\{\mathbf{x}_1^{j-1},\mathbf{x}_2^{j-1},\ldots,\mathbf{x}_m^{j-1}\})$ denotes the concatenated context representations of the $(j{-}1)$-th layer in the answering model.
    We write $\mathbf{E}$ for the last layer output of the prompting model.
    
    \item \textbf{Prompting answering model.} 
    Finally, the generated prompt $\mathbf{E}^j$ is prepended to the attention layer of the encoder $\operatorname{Enc}$ of the answering model as in Eq.~\ref{eq:prompt-tuning}.
    Meanwhile, the decoder $\operatorname{Dec}$ of answering model attends to $\operatorname{Cat}(\mathbf{E},\mathbf{X})$ and generates the target answer $a_t$:
    \begin{equation}
        p(a_{t}|\phi(a_{<t}),q,\mathcal{C})=\operatorname{Dec}(\operatorname{Cat}(\mathbf{E},\mathbf{X})).
    \end{equation}
\end{enumerate}
%

\noindent%
Capturing long-range dependencies among derived answers via a retrospective prompting mechanism enables the answering model to compose new contents grounding on what has already been devised, and thus the model is able to strike a good relevance-diversity balance for answering ambiguous questions.

\subsection{Answering ambiguous questions via iterative prompting}
\label{sec:iterative-prompt}


Given the input context, i.e., the question and retrieved evidential passages, AmbigPrompt iteratively performs attention operations over the input context and the generated answers,
 addressing the answer generation and prompt construction interactively.
The key is to pass the attention activations between the prompting model and answering model so that they can inspect each other's internal states and make harmonious predictions.
Specifically, we start from an empty answer set and progressively append newly generated answers to it.
As depicted in Figure~\ref{fig:model}, in each iteration, we first use the previously generated answer sequence to obtain the introspective prompts, and then interwoven the resultant prompting vectors into the answering model to predict the next answer.
Our algorithm terminates if the model reaches the \code{[EOI]} token.


\subsection{Optimization} \label{sec:optimization}
To enhance the pre-training model towards multi-answer QA, one straightforward approach is to leverage a question-answering dataset such as NQ~\citep{Kwiatkowski2019NaturalQA} for domain-adaptive pre-training~\citep{Min2021JointPR}.
However, the effectiveness of such a trivial approach is limited to the inherent defect of the one-pass prediction process; 
that is, the lack of the modeling capability of the interactions between answer generation and answer perception, which is critical to achieving superior performance in multi-QA scenarios.
To explicitly align the pre-training objective to task-specific preferences, we further propose to conduct task-adaptive post-pretraining on pseudo multi-answer QA dataset, and then finetune the proposed model using the task data.


\paragraph{Task-adaptive post-pretraining.} 
We first pre-train the model on NQ, in which only one answer $\mathcal{A}=\{a_1\}$ is labeled for each question $q$.
To explicitly characterize the pretraining stage as the efforts for finding \textit{which part of preceding answers to interact with regarding the input context}, 
we construct the pseudo multi-answer dataset $\hat{\mathcal{A}}$ for post-pretraining the proposed framework to mimic the iterative question answering process.
Specifically, we first train an \emph{auxiliary reader} $g(a|q,c_i)$, which learns to find an answer from the passage $c_i$ given a question $q$. 
Then, we use this auxiliary reader to generate a pseudo answer for each retrieved passage in $\mathcal{C}$:
\begin{equation}
    \hat{\mathcal{A}}=\{\hat{a} \mid \forall i \in [1,m], \hat{a}\sim g(a|q,c_i)\},
\end{equation}
where $\hat{\mathcal{A}}$ denotes the pseudo-answer set of $q$.

Then, we aggregate the generated answers to construct the previously known answers $a_{<t}$ in Eq.~\ref{eq:iter}.
In particular, we randomly sample $t$ answers from $\hat{\mathcal{A}}$ and filter out those that are equivalent to the ground-truth answer $a_1$; we denote the sampled set as $\hat{a}_{<t}$.
With the pseudo answers, we define the post-pretraining objective as:
\begin{equation}
    \mathcal{L}_{\text{Pre}} = -\log p(a_1|\phi(\hat{a}_{<t}),q,\mathcal{C}),
\end{equation}
where the number of answers in $\hat{a}_{<t}$, i.e., $t$, is sampled from a Bernoulli distribution.

\paragraph{Prompt-based fine-tuning.} 
We fine-tune the pre-trained model on downstream multi-answer QA datasets.
Specifically, in multi-answer QA, $n$ answers $\mathcal{A}=\{a_1,a_2,\ldots,a_n\}$ corresponding to a question $q$ are provided.
The model is tuned by the following objective:
\begin{equation}
    \mathcal{L}_{\text{FT}} = -\log p(a_t|\phi(a_{<t}),q,\mathcal{C}),
\end{equation}
where $t\in [1,n]$ is sampled from a Bernoulli distribution.
Since $\mathcal{A}$ is unordered, we shuffle $\mathcal{A}$ when constructing the $a_{<t}$ and $a_t$ to improve the robustness.
Besides, we explicitly optimize the model to generate \code{[EOI]} to stop the iteration.
Specifically, we define a parameter $\alpha \sim \mathcal{U}(0,1)$ and a thres\-hold $\lambda$, which controls the propensity of generating \code{[EOI]}.
If $\alpha < \lambda$, we replace the $a_t$ and $a_{<t}$ as \code{[EOI]} and $\mathcal{A}$, respectively.


\section{Experimental Setup}

\subsection{Datasets}
We evaluate AmbigPrompt on the AmbigQA~\citep{Min2020AmbigQAAA} and WebQSP~\citep{Yih2016TheVO} datasets.
\textbf{AmbigQA}:
AmbigQA is constructed to address the ambiguity of questions in open-domain QA. 
It samples 14,042 questions from NQ-Open~\citep{Kwiatkowski2019NaturalQA}, 
and asks annotators to search for, navigate and read multiple Wikipedia pages to find as many answers as possible.
%
\textbf{WebQSP}:
WebQSP consists of questions from Google Suggest API, originally from \citet{Berant2013SemanticPO}. The answer is a set of distinct entities in Freebase; we use the modified versions by \citet{Min2021JointPR}, which recasts WebQSP as textual question answering based on Wikipedia.


The statistical details of these two datasets and NQ are shown in Table~\ref{table:data}. 
\begin{table}[!t]
\centering \small
\resizebox{\columnwidth}{!}{
\begin{tabular}{@{}l rrr cc@{}}

\toprule
& \textbf{NQ} & \textbf{AmbigQA} & \textbf{WebQSP} \\
\midrule

Training size & 307,373 & 10,036 & 2,752 \\
Validation size  & 6,000 & 2,002 &  245\\
Test size & 6,000 & 2,004 &  1,582 \\
Mean \# Answers & 1.0 & 2.2 & 22.6 \\
Median \# Answers & 1.0 & 2.0 & 1.0 \\

\bottomrule
\end{tabular}
}
\caption{Data statistics of NQ, AmbigQA, and WebQSP.}
\label{table:data}
\vspace{-1em}
\end{table}

\subsection{Evaluation metrics}
Following previous studies~\citep{Min2020AmbigQAAA}, we adopt F1 as the evaluation metric, which measures the precision and recall between the ground-truth answers and the predicted answers.
The test set is further divided into two subsets: \emph{full} and \emph{multi}.
The \emph{full} subset evaluates the model on all the questions in the test set, while the \emph{multi} subset evaluates the model on the questions with multiple answers (i.e., $n>1$).
To assess the computational efficiency of various approaches, we also report the number of parameters, average latency, and peak memory usage during model inference.
All the models are tested on the same device.
We estimate the latency and memory usage of those baselines without public code using randomly initialized models since these metrics are independent of their parameters given a fixed number of encoded tokens and decoding length.

\begin{table*}[!t]
\centering \small
\setlength\tabcolsep{5pt}
\begin{tabular}{@{}l rr  rr  rr  rr  rr@{}}

\toprule

\multirow{2}{*}{\textbf{Methods}}
& \multicolumn{2}{c}{\textbf{AmbigQA}}
& \multicolumn{2}{c}{\textbf{WebQSP}}
& \multicolumn{2}{c}{\multirow{2}{*}{\textbf{\#Params}}} 
& \multicolumn{2}{c}{\multirow{2}{*}{\textbf{Latency}}}
&\multicolumn{2}{c}{\multirow{2}{*}{\textbf{Memory}}}
\\
\cmidrule(r){2-3}
\cmidrule{4-5}

&  Full & Multi & Full & Multi &&&&&
\\

\midrule
\multicolumn{10}{@{}l}{\emph{High-capacity baselines}}\\
JPR$^\dagger$~\citep{Min2021JointPR} & 48.5 & 37.6 & 53.1 & 47.2 & 3B & $8.7\times$ & 0.88s & $2.3\times$  & 14GB & $3.5\times$\\


RECTIFY$^\dagger$~\citep{Shao2021AnsweringOM} & 52.1 & 41.6 & 55.8 & 48.8 & 6B & $17.4\times$ & 19.72s & $51.3\times$ & 14GB & $3.5\times$ \\

\midrule

\multicolumn{10}{@{}l}{\emph{Comparable low-capacity baselines}}\\
DPR~\citep{Karpukhin2020DensePR} & 38.9 & 29.9 & 44.7 & 35.5 & 345M & $1.0\times$ & 0.37s & $1.0\times$ & 4GB & $1.0\times$ \\

SpanSeqGen~\citep{Min2020AmbigQAAA} & 39.7 & 29.3 & 48.8 & 36.1 & 400M & $1.2\times$ & ~~0.49s & $1.3\times$ & 5GB & $1.3\times$ \\

FiD-Base~\citep{Izacard2021LeveragingPR} & 45.5 & 35.8 & 52.6 & 46.3 & 220M & $0.6\times$ & 0.38s & $1.0\times$ & 4GB & $1.0\times$ \\

Refuel$^\dagger$~\citep{Gao2021AnsweringAQ} & 48.3 & 37.3 & -- & -- & 400M & $1.2\times$ & 22.19s & $58.6\times$  & 8GB & $2.0\times$ \\

\midrule

\textbf{AmbigPrompt} & \textbf{48.7} & \textbf{38.8} & \textbf{53.2} & \textbf{47.9} & 220M & $0.6\times$ & 0.68s & $1.8\times$ & 4GB & $1.0\times$  \\

\bottomrule
\end{tabular}
\caption{Results on AmbigQA dev and WebQSP test in terms of effectiveness and efficiency. Full and Multi denote the full set and multi-answer set, respectively. The reported value is F1. Methods with $^\dagger$ have no publicly available codes; therefore, we estimate the latency and memory footprint with randomly initialized parameters. 
We divide baselines into two groups: 
(i) \emph{high-capacity baselines} that use significantly larger models than AmbigPrompt, and 
(ii) \emph{comparable low-capacity baselines} that use a low-capacity model like AmbigPrompt and can be reasonably compared with AmbigPrompt.
\textbf{Boldface} indicates best performance among comparable baselines.} \label{table:ambigqa}
\end{table*}

\subsection{Baselines}
The following models are adopted as baselines:
\textbf{DPR}~\citep{Karpukhin2020DensePR}:
A dual-encoder is trained using contrastive loss for passage retrieval, and a BERT-based reader is used for answer extraction.
\textbf{SpanSeqGen}~\citep{Min2020AmbigQAAA}:
DPR reranks the passages, and a BART-based generator is used for answer generation. 
\textbf{FiD}~\citep{Izacard2021LeveragingPR}:
The retrieved passages are encoded by a T5 encoder independently, and the representations are then concatenated and fed into the T5 Decoder to generate answers.
\textbf{Refuel}~\citep{Gao2021AnsweringAQ}:
A question disambiguation module is proposed to generate disambiguated questions. The disambiguated questions are then used to find more answers. 
%
\textbf{JPR}~\citep{Min2021JointPR}:
JPR is a passage reranker that reranks the passages using an autoregressive model.
With the additional reranking stage, JPR selects ten diverse passages from 100 retrieved passages and uses a T5-3B FiD answering model to compose answers in one pass.
\textbf{RECTIFY}~\citep{Shao2021AnsweringOM}:
RECTIFY proposes the recall-then-verify framework, which separates the reasoning process of each answer. An answering model operates on each passage to recall surplus answers. Then, a sophisticated verifier based on T5-3B FiD verifies each answer with an aggregation module.

We divide the baseline models into two categories depending on the number of parameters of the models: 
(i) \emph{high-capacity baselines} that use large models with billions of parameters, while requiring more computational resources and memory; 
(ii) \emph{comparable low-capacity baselines} that use low-capacity models with a similar number of parameters and computational effort as AmbigPrompt, which can be reasonably compared with AmbigPrompt.

\subsection{Implementation details}
We choose T5-Base~\citep{Raffel2020ExploringTL} as the backbone of the answering model.
Regarding the passage retrieval model, we fine-tune the pre-trained model from \citet{Gao2021UnsupervisedCA} on the NQ dataset (See Appendix~\ref{sec:retrieval} for details).
The retrieval corpus is the English Wikipedia on 12/20/2018, and the documents are split into chunks with 100 words following \citet{Karpukhin2020DensePR}.
We set $m{=}100$, $\lambda{=}0.5$, the batch size to $32$, and the model is trained using the AdamW optimizer~\citep{Loshchilov2017DecoupledWD} with a constant learning rate of $5e{-}5$.
We train the model up to 5k steps on on 4 V100-16G GPUs and choose the hyperparameters and checkpoints on the validation set.\footnote{Since we test on the AmbigQA dev set, we slice about 1k examples in the AmbigQA training set as the validation set.}


\section{Experimental Results}

\subsection{Main results}
Table \ref{table:ambigqa} reports the evaluation results on AmbigQA and WebQSP.
Based on the results, we have three main observations.

First, AmbigPrompt achieves comparable performance to the state-of-the-art.
Specifically, AmbigPrompt obtains 48.7 F1 on the \emph{full} test set and 38.8 F1 on the \emph{multi} test set, which exceeds all baselines except RECTIFY.
The improvements are particularly significant on the \emph{multi} test set; AmbigPrompt improves 1.2\% over JPR and 1.5\% over Refuel.
Besides, compared with FiD, which concatenates all the answers in $\mathcal{A}$ with \code{[SEP]} and generates them in one pass, 
the proposed method, which benefits from the iterative design and answer-conditional prompting mechanism, achieves 3\% and 5\% improvements on \emph{full} and \emph{multi} of AmbigQA.
Similar results can also be observed on WebQSP.

Second, AmbigPrompt uses fewer resources compared to previous high-capacity models.
AmbigPrompt uses a lightweight model with 220M parameters.
Still, AmbigPrompt achieves superior performance compared to the high-capacity models, e.g., JPR, that use 3B parameters.
The state-of-the-art model RECTIFY uses 6B parameters (3B for the answering model and 3B for the verifier), which is $27 \times$ as much as ours, significantly increasing the training and inference overhead.
Similar results are witnessed in terms of latency.
In particular, RECTIFY is $29 \times$ slower than our model due to the heavy design of the answering model and verifier.
Refuel's iterative passage retrieval and clarifying question generation procedure results in a $32.6 \times$ latency compared with our approach.
Finally, the comparison of peak memory usage also confirms our approach's lightweight nature.
The lightweight design allows our approach to be adapted to academically accessible devices and reduces the carbon footprint for model training and deployment.

\begin{figure}[t]
 \centering
 \includegraphics[width=1\columnwidth]{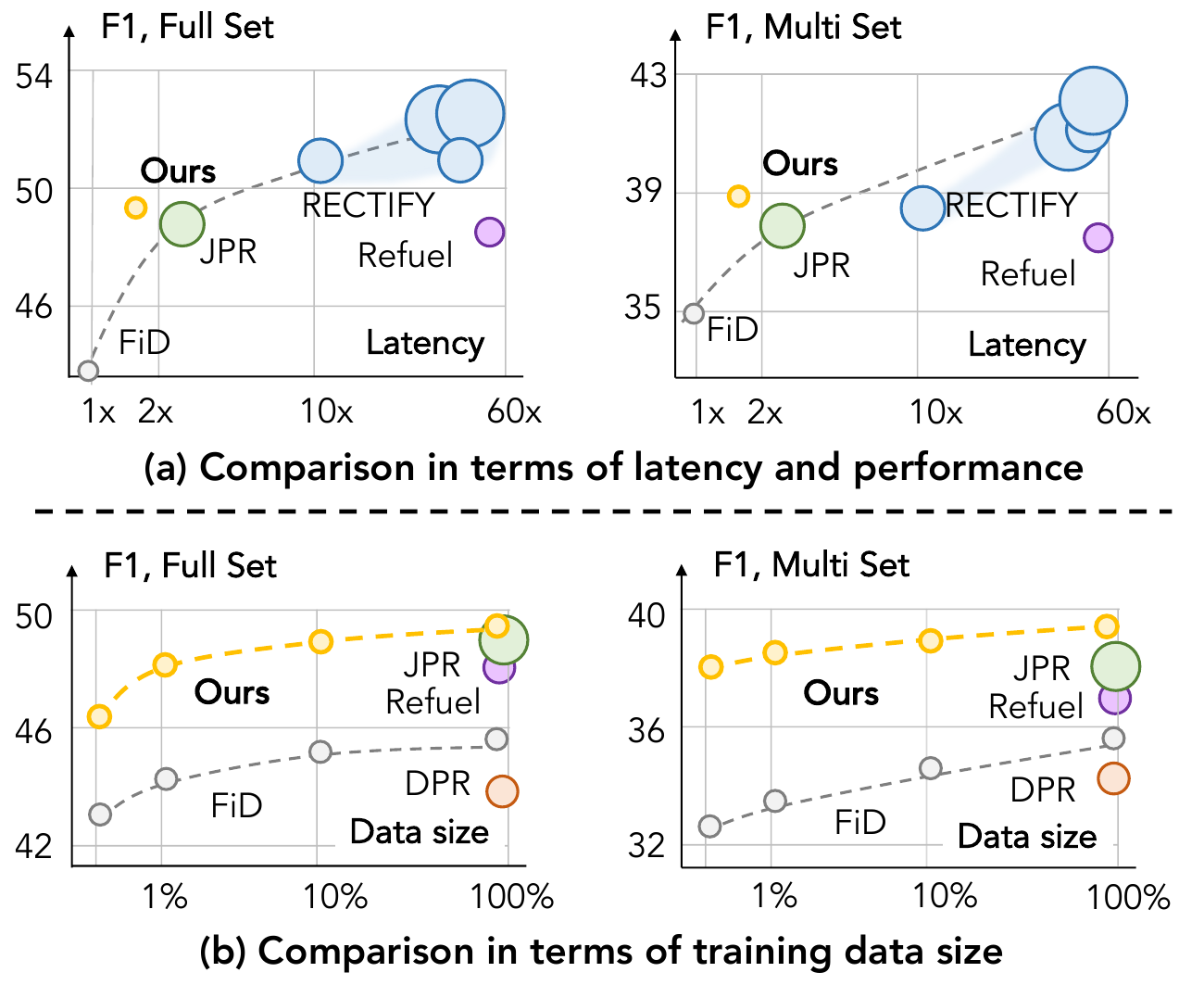}
 \caption{(a) Latency (in log scale) versus performance (F1) on AmbigQA full dev and multi dev. The size of the circle indicates the number of parameters of these models. (b) Dataset size (in \%) versus performance (F1) on AmbigQA full dev and multi dev.}
 \label{fig:speed}
\end{figure}

Third, we find that AmbigPrompt achieves a better resource-performance balance.
In Figure~\ref{fig:speed}~(a), we display the existing methods under the speed-performance coordinate system.
Note that we place RECTIFY with different sizes (i.e., latency) on the diagram according to \citet{Shao2021AnsweringOM}.
AmbigPrompt improves the optimal latency-performance curve (the dashed lines), especially on the multi-answer test set, demonstrating the effectiveness of our approach in answering ambiguous questions.

\subsection{Low-resource setting}
Figure~\ref{fig:speed}~(b) shows the results under different training data sizes to investigate the effectiveness of the proposed method in the low-resource setting.
The proposed method achieves favorable results for different data sizes. 
Remarkably, AmbigPrompt achieves promising performance with little data, surpassing the fully supervised high-capacity model JPR on a multi-answer test set.
This result suggests that the proposed prompting mechanism can better elicit the capabilities of the pre-trained model and effectively adapt the model trained on single-answer QA data to multi-answer scenarios.


\begin{table}[!t]
\centering \small
\resizebox{\columnwidth}{!}{
\begin{tabular}{@{}l cc cc@{}}

\toprule

\multirow{2.5}{*}{\textbf{Methods}} & \multicolumn{2}{c}{\textbf{AmbigQA}}
& \multicolumn{2}{c}{\textbf{WebQSP}} 
\\
\cmidrule(lr){2-3} \cmidrule(lr){4-5} 

& Full & Multi & Full & Multi\\
\midrule
AmbigPrompt & 48.7 & 38.8 & 53.3 & 46.7 \\
\midrule
- w/o task-adaptive pre-training & 42.8 & 32.7 & 42.5 & 38.7\\
- w/o prompting model & 46.0 & 34.3 & 49.7 & 44.6\\
- w/o interleaving prompting  & 47.8 & 36.9 & 50.9 & 45.4 \\
\bottomrule
\end{tabular}
}
\caption{Ablation study. The base model is compared with several ablative variants on two datasets.}\label{table:ablation}
\end{table}

\subsection{Ablation study}
To understand the contribution of each component of AmbigPrompt, we conduct an ablation study.
The results are listed in Table~\ref{table:ablation}.
The compared variants and the findings are:

\header{W/o task-adaptive pre-training}
The models are trained only on multi-QA data with $\mathcal{L}_{PT}$.
A notable performance decline can be seen.
This observation suggests that task-adaptive pre-training is an important contributor to the model's performance since the size of multi-answer QA data is small.

\header{W/o prompting model}
We remove the prompting model in this variant and instantiate the learnable prompt vector to each step $t$ separately, like \citet{Liu2021PretrainPA}.
The performance drops by about 3\% and 4\% on the two datasets, respectively.
The results verify the effectiveness of the proposed answer-conditional prompting mechanism.

\header{W/o interleaving prompting}
We remove the interaction mechanism between the prompting model and answering model, i.e., the FiD encoder encodes the $e$ and $X$ independently without cross-attention.
The results drop by about 2\% and 2\% on two datasets, respectively,
which reveals that enabling the answering model to generate new answers conditioned on the introspective prompts effectively improves the model's performance.

\subsection{Analytical experiments}
\begin{figure}[!t]
 \centering
 \includegraphics[width=1\columnwidth]{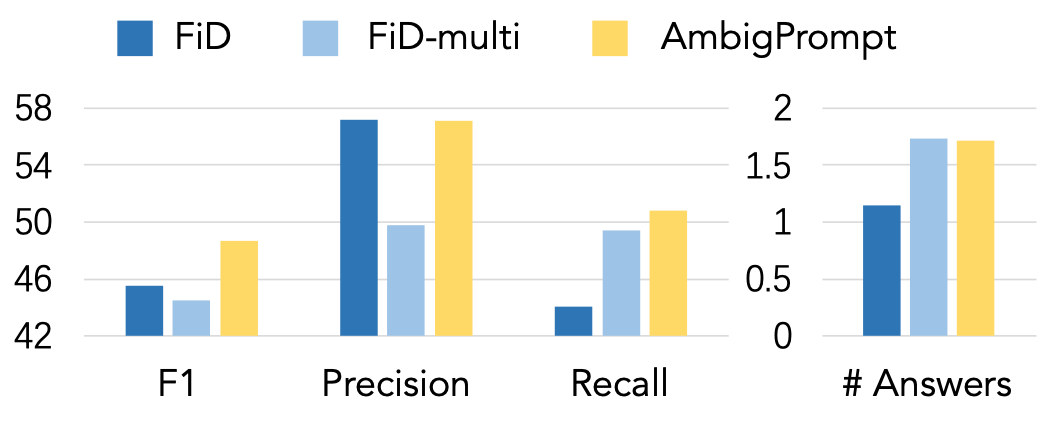}
 \vspace*{-8mm}
 \caption{The F1, Precision, Recall, and the average number of answers (\#Answers) of AmbigPrompt and FiD model variants on AmbigQA dev.}\label{fig:recall}
\end{figure}

Conceptually, our proposed framework AmbigPrompt equips the FiD model with the ability to progressively compose the answers using retrospective prompts, i.e., iterative prompt learning.
To further analyze the capability of such an iterative prompt learning approach in managing the relevance-diversity trade-off, we present the F1, precision, recall, and average answer numbers of AmbigPrompt and FiD model variants in Figure~\ref{fig:recall}.
In particular, FiD-multi denotes a variant of FiD in which we reduce the generation probability of the end-of-sequence token \code{</s>} to ensure that the number of generated answers is approximately the same as AmbigPrompt.
We see that FiD-multi obtains comparable recall but gets significantly lower precision.
In contrast, AmbigPrompt generates more answers than FiD without sacrificing precision, indicating that the designed iterative prompting mechanism induces the model with a superior ability to manage the trade-off between relevancy and diversity for ambiguous question answering.

\section{Related work}

\subsection{Ambiguous question answering}\label{sec:open-qa}
In open-domain QA, given a question about any topic, the model finds the answer from a large knowledge corpus~\citep{Chen2017ReadingWT}.
Typically, a retrieval model and an answering model are employed.
The two modules can be trained separately~\citep{Karpukhin2020DensePR,Izacard2021LeveragingPR,Qu2021RocketQAAO} or jointly~\citep{Lee2021YouON,Lewis2020RetrievalAugmentedGF,Izacard2021DistillingKF}.
Ambiguity is inherent to open-domain QA; especially when exploring new topics, it can be difficult to ask questions that have a single, unambiguous answer~\citep{Min2020AmbigQAAA,Rubin2022QAMPARIA}.
\citet{Min2020AmbigQAAA} identify the challenge of \emph{multi-answer QA} and collect the dataset AmbigQA.
Based on that, \citet{Min2021JointPR} propose an autoregressive passage reranking model JPR, which reranks the top-retrieved passages and improves their diversity.
\citet{Gao2021AnsweringAQ} propose a round-trip prediction approach, where clarification questions are generated and fed back into the model to find more answers.
\citet{Shao2021AnsweringOM} propose a recall-and-verify framework, where surplus answers are generated first, and a verifier model then determines each candidate answer.
Compared with existing methods, we propose a lightweight yet effective approach to answering ambiguous questions by iterative prompting.

\subsection{Prompt-based learning}\label{sec:prompt-based-learning}
Prompt-based learning has received much attention recently~\citep{Liu2021PretrainPA}. 
Existing studies on prompt-based learning mainly focus on discrete and continuous prompts.
The former designs text-based prompts~\citep{Jiang2020HowCW,Gao2021MakingPL,Schick2021ItsNJ}, while the latter prepend a learnable prompt vector to word embeddings~\citep{Lester2021ThePO,Liu2021GPTUT} or attention layers~\citep{Li2021PrefixTuningOC,Liu2021PTuningVP}.
Prompt-based learning has demonstrated advantages in low-parameter tuning~\citep{He2021TowardsAU} and few-shot/zero-shot performance~\citep{Brown2020LanguageMA,Wei2021FinetunedLM}.
We propose an iterative prompting method for multi-answer QA based on answer-conditional continuous prompts.

\subsection{Iterative generation} \label{sec:iterative-gen}
Iterative generation (a.k.a.\ progressive generation) aims to decompose a challenging generation task into multiple steps and progressively produce the target sequence.
Iterative generation has been applied to the tasks of machine translation~\citep{Lee2018DeterministicNN}, controllable text generation~\citep{Casas2020SyntaxdrivenIE,Zhang2020POINTERCP}, storytelling~\citep{Hua2020PAIRPA,Tan2021ProgressiveGO}, data-to-text~\citep{Kasner2020DatatoTextGW}, etc.
Recently, \citet{Wang2022ShepherdPL} introduced an iterative prompting framework to progressively elicit knowledge from language models for commonsense reasoning and multi-hop question answering tasks~\citep{Qi2019AnsweringCO,Xiong2021AnsweringCO}.
Compared to existing work, we propose an answer-conditional prompting model and an effective task-specific pre-training scheme for multi-answer QA.

\section{Conclusions}
In this paper, we have proposed AmbigPrompt for multi-answer QA.
AmbigPrompt is a simple yet effective model that answers ambiguous questions by iterative prompting.
We have proposed an answer-conditional prompting model for prompt generation, and a task-adaptive post-pretraining scheme for model training.
Extensive experiments suggest that AmbigPrompt achieves comparable performance as high-capacity models and achieves the best results in a low-resource setting.

\section*{Limitations}
The limitations of this paper include the absence of experiments on large language models.
Previous studies have shown that using high-capacity pre-trained language models can significantly improve the accuracy of answers but also entails an increase in computational overhead. 
Due to (academic) limitations of computational resources, this paper employs a low-capacity T5 model for experiments.
Our experiments have suggested that the proposed iterative prompting method that works with the low-capacity model can achieve comparable results with baseline methods equipping with large models.

In future work, we would like to scale up the proposed model to improve the model's performance.
\changed{
Recent research on large language models (LLMs) has shown that they can learn from few examples and reason well. We believe that it is worth exploring ways to enhance the prompting of LLMs to improve their completeness when responding to ambiguous questions and reduce model hallucination in generation~\citep{OpenAI2023GPT4TR,LLMSurvey,Sun2023ContrastiveLR}.
Another direction worth exploring in the future is the application in low-resource scenarios, such as low-resource languages.
Low-resources in our study are characterized by limited multi-answer-QA annotations, which aims to examine how data size impacts model performance. 
Other low-resource languages may behave differently with less training data and large models~\citep{Xue2020mT5AM,Sun2021ConversationsPB}.}
Besides, we would like to explore more effective prompting methods, such as chain-of-thought prompting~\citep{Wei2022ChainOT}.

\section*{Ethics Statement}
The paper has proposed a question-answering model, which is intended to answer factoid open-domain questions. 
The model-predicted answers still have a considerable amount of misinformation.
Besides, the proposed models rely on pre-trained question-answering models, which are trained on large-scale web data that is known to contain biased or discriminatory content.

\section*{Acknowledgements}
This work was supported by the National Key R\&D
Program of China with grant No. 2020YFB1406704,
the Natural Science Foundation of China (62272274, 61972234, 62072279, 62102234, 62202271), 
the Natural Science Foundation of Shandong Province
(ZR2022QF004), 
the Key Scientific and Technological Innovation Program of Shandong Province (2019JZZY010129), 
the Fundamental Research Funds of Shandong University, 
the Hybrid Intelligence Center, a 10-year program funded by the Dutch Ministry of Education, Culture and Science through the Netherlands Organization for Scientific Research, \url{https://hybrid-intelligence-centre.nl}.

All content represents the opinion of the authors, which is not necessarily shared or endorsed by their respective employers and/or sponsors.




\bibliography{anthology,custom}
\bibliographystyle{acl_natbib}

\clearpage
\begin{appendices}

\appendix

\section{Results on NQ}
Table~\ref{table:nq} lists the exact match (EM) score of the baselines and AmbigPrompt on single-answer QA benchmark, NQ-Open test.
We see that the high-capacity models (e.g., JPR), which benefit from large language models like T5-3B, achieve better EM score.
However, in the multi-answer QA task, the models need to focus not only on the precision of answers, but also on the diversity of answers (i.e., recall rate).
In AmbigQA, we can see that the proposed model outperforms JPR, indicating its superior ability to recall multiple feasible answers.

\begin{table}[h]
\centering \small
\setlength\tabcolsep{6pt}
\begin{tabular}{@{}l cc c@{}}

\toprule

Method & \#Params  & EM \\
\midrule
JPR~\citep{Min2021JointPR} & 3B & 54.5\\
RECTIFY~\citep{Shao2021AnsweringOM} & 6B & 54.8\\
\midrule
DPR~\citep{Karpukhin2020DensePR} & 345M & 41.5 \\
SpnSeqGen~\citep{Min2020AmbigQAAA} & 400M & 45.0\\
FiD-Base~\citep{Izacard2021LeveragingPR}  & 220M & 48.2\\
FiD-Large~\citep{Izacard2021LeveragingPR}  & 700M & 51.4\\
\citet{Izacard2020DistillingKF} & 220M & 49.6\\
Refuel~\citep{Gao2021AnsweringAQ} & 400M & 48.9\\
\midrule
AmbigPrompt & 220M & 49.2\\
\bottomrule
\end{tabular}
\caption{Model size and EM score on NQ test.}
\label{table:nq}
\end{table}

\section{Zero-shot evaluation on AmbigQA}
We also test the proposed model and baselines on AmbigQA in zero-shot setting following \citet{Min2020AmbigQAAA}.
In zero-shot evaluation, the models are trained using partial supervision only (i.e., single-answer NQ-Open~\citep{Kwiatkowski2019NaturalQA}), and are evaluated on multi-answer data AmbigQA.
This setting provides a practical application where only single-answer datasets are available.
Note that the zero-shot evaluation on AmbigQA allows the model to tune some hyper-parameters (e.g., threshold of generation probability~\citep{Min2020AmbigQAAA}) using development data, which may make the setting not zero-shot in the strictest sense.

The compared models are (1) DPR and SpanSeqGen, in which the models trained on NQ-Open are adopted to predict multiple answers via a thresholding strategy~\citep{Min2020AmbigQAAA}.
(2) FiD with various decoding methods, in which FiD trained on NQ-Open produces multiple answers through (a) Nucleus sampling with $\{p{=}0.8, t{=}0.8\}$; (b) Top-k sampling with $\{k{=}40, t{=}0.8\}$; and (c) Diverse beam search with $\{b{=}3, t{=}0.8, \textit{diversity\_penalty}{=}0.5\}$.
We also evaluate FiD with greedy decoding that generates one answer for each question as the default setting of FiD.
(3) AmbigPrompt, in which the FiD answering model prompted by our proposed answer-conditional prompting model is trained on NQ-Open with our task-adaptive post-pretraining method and produces multiple answers through iterative prompting.

The results are listed in Table~\ref{table:zero-shot}.
FiD series outperform DPR and SpanSeqGen as they utilize more passages that potentially cover more feasible answers.
FiD with nucleus sampling obtains the best results among different decoding methods.
AmbigPrompt achieves the best zero-shot performance on AmbigQA and also outperforms high-capacity supervised baselines JPR on the multi-answer subset.

\begin{table}[h]
\centering\small
\setlength\tabcolsep{10pt}
\begin{tabular}{l cc}

\toprule

Methods & Full & Multi \\
\midrule
DPR & 35.2 & 26.5 \\
SpanSeqGen & 36.4 & 24.8 \\
\midrule
FiD & 43.7 & 33.5 \\
~- nucleus sampling & 45.7 & 36.7 \\
~- top-k sampling & 42.6 & 34.7 \\
~- diverse beam search & 45.2 & 36.1\\
\midrule
\textbf{AmbigPrompt} & \textbf{46.5} & \textbf{37.9}\\

\bottomrule
\end{tabular}

\caption{Zero-shot evaluation results on AmbigQA.}\label{table:zero-shot}
\end{table}

\section{Retrieval results}\label{sec:retrieval}
We train the dense retrieval model on NQ-Open using in-batch negatives with batch size 64.
The retrieval model is initialized from CoCondenser~\citep{Gao2021UnsupervisedCA}.
Our retrieval corpus is the English Wikipedia from 12/20/2018. 
Table~\ref{table:retrieval} lists the retrieval results on NQ-Open and AmbigQA.
In NQ-Open, we use Recall@k (R@k for short) as the metric, which considers retrieval to be successful if at least one answer is included in the top-k ranked passages.
In AmbigQA, we use MRecall@k (MR@k for short) as the metric, which considers retrieval to be successful if all answers or at least k answers in the answer set $\mathcal{A}$ are covered by the top-k ranked passages.
From the results, we see that our retrieval model achieves comparable results against baseline retrieval models, but underperforms reranking models such as KPR and MonoT5.

\begin{table}[h]
\centering \small
\setlength\tabcolsep{5pt}
\begin{tabular}{l cccc}

\toprule

NQ-Open & R@1  & R@5 & R@10 & R@100 \\
\midrule
DPR & 43.1 & 68.5 & 76.4 & 87.9\\
RECTIFY & - & 73.8 & - & 89.3 \\
Ours & 50.9 & 72.2 & 78.2 & 88.2\\

\midrule

AmbigQA & MR@1  & MR@5 & MR@10 & - \\
\midrule

DPR & -& 55.2 & 59.3 & -\\
RECTIFY & - & 53.2 & 60.0 & -\\
MonoT5$^\dagger$ & - & 63.4 & 65.8 & -\\
JPR$^\dagger$ & - & 64.8 & 67.1 & -\\
Ours & 61.7 & 56.4 & 62.6 & -\\

\bottomrule
\end{tabular}
\caption{Retrieval results on NQ-Open test and AmbigQA dev. $^\dagger$ denotes reranking model.}
\label{table:retrieval}
\end{table}

\section{Case study}
We present some examples in Table~\ref{table:case1} and Table~\ref{table:case2}.





\begin{table}[!h]
\small \centering
\setlength\tabcolsep{4pt}

\begin{tabular}{@{}rp{6.5cm}@{}}

\toprule
\textbf{Question} &{Who holds the record for most passing yards in a season?}\\
\midrule
\multirow{2}{*}{\textbf{Passages}}
& Associated Press NFL Offensive Player of the Year Award | Marino's 5,084 yards stood as the record for 27 years before being broken by Drew Brees in 2011, who won that season's award. In turn, 2013 winner Peyton Manning set league single-season records for passing yards (5,477) and passing touchdowns (55). [...]\\
\midrule
\textbf{FiD} &{drew brees, peyton manning}\\
\textbf{Ours} &{drew brees, dan marino, peyton manning}\\
\textbf{Human} &{Peyton Manning, Drew Brees, Dan Marino}\\
\bottomrule
\end{tabular}
\caption{An example on AmbigQA dev shows that the proposed method AmbigPrompt finds all valid answers.} \label{table:case1}
\end{table}


\begin{table}[!h]
\small \centering
\setlength\tabcolsep{4pt}

\begin{tabular}{@{}rp{6.5cm}@{}}

\toprule
\textbf{Question} &{Who was the bond girl in you only live twice?}\\
\midrule
\multirow{2}{*}{\textbf{Passages}}
& Severine | She had also categorized Aki and Kissy Suzuki, both from "You Only Live Twice" (1967), as falling into this trope. She supported this assessment by pointing to the character\'s lack of agency and impact on "Skyfall"\'s main narrative, and summed up Sévérine as "one of the most disempowered, pitiful, and tragic women in the Bond film franchise". [...]\\
& You Only Live Twice (film) | Sean Connery's then-wife Diane Cilento performed the swimming scenes for at least five Japanese actresses, including Mie Hama. Martial arts expert Donn F. Draeger provided martial arts training, [...]
\\
\midrule
\textbf{FiD} &{akiko wakabayashi}\\
\textbf{Ours} &{aki, kissy suzuki, yasuko nagazumi, akiko wakabayashi}\\
\textbf{Human} &{Aki, Akiko Wakabayashi, Kissy Suzuki, Mie Hama}\\
\bottomrule
\end{tabular}
\caption{An example on AmbigQA dev shows that AmbigPrompt finds more valid answers than FiD.} \label{table:case2}
\end{table}

\end{appendices}

\end{document}